\documentclass[]{PHMSociety}

\usepackage{comment}
\usepackage{graphicx}
\usepackage{amsmath}
\usepackage{amsfonts}
\usepackage{amssymb}
\usepackage{enumerate}
\usepackage{algorithm2e}
\usepackage{booktabs}
\usepackage{url}
\usepackage{multirow}
\usepackage{xcolor}
\usepackage{makecell}

\newcommand{\eq}[1]{(\ref{#1})} 

\newcommand{\expect}{{\mathbb E}}

\begin{document}

\title{Data Augmentation of Multivariate Sensor Time Series using Autoregressive Models and Application to Failure Prognostics}
\author{%
	Douglas Baptista de Souza and Bruno Paes Leao \authorNumber{} \\ 
}

\address{
	\affiliation{PAPER PREPRINT}{ }{ 
		{\email{}}
		} 
	\tabularnewline 
	\affiliation{ }{}{ 
		{\email{ }}
		} 
}

\maketitle


\begin{abstract}
This work presents a novel data augmentation solution for non-stationary multivariate time series and its application to failure prognostics. The method extends previous work from the authors which is based on time-varying autoregressive processes. It can be employed to extract key information from a limited number of samples and generate new synthetic samples in a way that potentially improves the performance of PHM solutions. This is especially valuable in situations of data scarcity which are very usual in PHM, especially for failure prognostics. The proposed approach is tested based on the CMAPSS dataset, commonly employed for prognostics experiments and benchmarks. An AutoML approach from PHM literature is employed for automating the design of the prognostics solution. The empirical evaluation provides evidence that the proposed method can substantially improve the performance of PHM solutions.
\end{abstract}

\section{Introduction}

PHM has been an active topic in research and solution development during the recent decades. The motivation is in associated with benefits such as reduced downtime, improved yield and safety which can be enabled by failure diagnostics and prognostics. Many related methods have been proposed over the years, moving from traditional reliability-based practices based on population statistics to advanced data-driven and physics-based solutions which can estimate current and future health states associated with specific assets and failure modes. Data-driven solutions benefit from the great advances recently achieved in the field of machine learning to produce accurate diagnostics and prognostics estimates. However, despite the great advances in PHM methods, sufficient historical data associated with the failure modes of interest is required to apply such methods in real world, and many times such data is not available \cite{biggio2020, kim2021}. The underlying reason may be simply that failures can be rare, and the justification of developing associated PHM solutions comes from the high impact associated with a potential occurrence. However, even in cases where related failure events have happened a reasonable number of times in the past, relevant sensor data associated with those historical events may not have been collected or, if collected, it may have quality issues or may not be associated with corresponding labels defining the actual failure modes.

Such challenges associated with the data naturally affect more directly data-driven methods, but physics-based methods are many times also impacted as failure mechanism models may be very complex and depend on availability of historical data for parameter tuning. Whenever enough data is not available, despite all the advances of PHM methods, the development of failure diagnostics or prognostics solution may result in poor performance or may not even be feasible. In such cases, the possibilities for PHM solution development are in general limited to anomaly detection which is not as prescriptive or actionable.

Given the potential limitations imposed by the lack of sufficient good quality data, data-centric approaches can be valuable for enabling application of PHM solutions in the real world \cite{leao2021, garan2022}. Data-centric in this context may be related to improving data collection and labeling or making the most out of available data. Data augmentation is one of the most promising mechanisms for achieving the latter, having gained increasing attention over the recent years in the contexts of both failure diagnostics \cite{matei2018, kwak2023} and prognostics \cite{liphmaugmentation2024, kim2020}, and resulting in improved performance in various PHM use cases \cite{limiteddatadiagdaug1, limiteddatadiagdaug2, limiteddatadiagdaug3, limiteddatadiagdaug4, limiteddatadiagdaug5, limiteddatadiagdaug6, limiteddatadiagdaug7}.

Previous work from the authors \cite{daug_tvar_phm_2023} comprised a novel method for augmentation of multivariate time series data based on time-varying autoregressive (TVAR) models. The goal was to extract information from scarce data and use it to create additional samples in a way that can improve the quality PHM solutions. The method was successfully employed to improve performance in failure diagnostics solutions. The work described here extends such methodology by proposing improvements to the TVAR models that make them more suitable for modeling multivariate sensor time series. The resulting enhanced solution is applied to the more challenging domain of failure prognostics. 

Empirical assessment of the proposed method is performed based on the C-MAPSS dataset, which is traditionally employed for benchmarking failure prognostics methods \cite{ramasso2014review}. In order to ensure that the assessment of the proposed data augmentation method is decoupled from a specific choice of prognostics approach, the automated machine learning (AutoML)-based solution described in \cite{kefalas2021automated} and available in github\footnote{https://github.com/MariosKef/automated-rul} was employed. The authors of the referred work proposed the creation of prognostics methods with a fully automated pipeline that includes the definition of the machine learning approach, hyperparameter tuning, and final performance metrics evaluation. They implemented the AutoML using Python library Tree-based Pipeline Optimization Tool (TPOT)\footnote{https://epistasislab.github.io/tpot2/} and performed experiments using the same C-MAPSS dataset employed here, achieving good results. Therefore, it presents favorable characteristics for testing data-centric prognostics methods.

The remaining of this paper is organized as follows: section \ref{sec:tvarmodel} presents the updated multivariate TVAR model; in section \ref{sec:data_augmentation} the proposed application of TVAR for data augmentation is described; experiments and results are presented and discussed in section \ref{sec:exp_study}; section \ref{sec:conclusion} is the conclusion. 

\section{The TVAR Model} \label{sec:tvarmodel}

Sensor time series are often non-stationary. Models to characterize such time series have been proposed in the past for different purposes (e.g., process modeling, forecasting, anomaly detection). Classical models such as Autoregressive (AR) and Autoregressive Moving Average (ARMA) can be employed in this context, but they require identifying and removing non-stationary components from the time series in advance \cite{kaynonstatdetect, nonstaticassp2012, nonstattrendtimemarginal, slownonstat, splstat, nonstatemdflorence}. In the past decades, models of the type Autoregressive Integrated Moving Average (ARIMA) became a popular choice in many applications \cite{booktimeseriesmodels}. Examples are the Seasonal ARIMA (SARIMA) \cite{sarima2017} and the Fractional ARIMA (FARIMA) models \cite{farima2}, which are useful when modeling time series exhibiting seasonality and long-memory behaviors, respectively. However, one disadvantage of such models is that the potential nonstationarity in the data should be of particular forms (e.g., seasonality), so that data transformations such as differencing can be employed.  

Recently, machine learning models based on deep learning and transformer architectures have excelled in tasks such as time series forecasting \cite{google_time_series, llm_time_series}, even when handling non-stationary data. However, these models have disadvantages in terms of the amount of data required for training and the lack of interpretability. Such approaches do not produce an analytical representation of the modeled process, which is an important feature in many engineering applications. In this work, we make use of a TVAR process to model multivariate sensor time series. This approach can be employed even when only few data samples of the modeled process are available, which is usually the case when performing failure prognostics. It also provides an analytical representation of the data and has been successfully employed to characterize real-world time series exhibiting non-stationary behaviors that cannot be modeled by classical AR and ARMA, as well as ARIMA-based models. Examples of non-stationary time series data that have been modeled by TVAR processes include electroencephalography (EEG) \cite{eegtvar}, acoustic signals \cite{tvaracousticsigs}, and fluctuations in electric networks \cite{tvarnonstatspl}).  More details on TVAR models can be found in classical references such as \cite{oldtvarref}, or more recent ones like \cite{tvarnonstatspl}.

For employing the TVAR model to create augmented multivariate time series, we consider the following assumptions:

A1) The time series are sampled equally over time. \\
A2) The time series data can be modelled by the same mutivariate, potentially non-stationary stochastic process.

The potential non-stationary behavior means that the properties of the stochastic process can vary over time. Similarly to \cite{daug_tvar_phm_2023}, the data augmentation adopted here follows from assumption A2, and consists of seeking a candidate stochastic process to characterize the time-series data. Realizations of this stochastic process are then employed as new synthetic samples of the corresponding time series. 

In \cite{tvarnonstatspl}, a TVAR model of first-order has been proposed to model non-stationary processes whose mean and covariance vary over time following an arbitrary scalar functional form defined by the user. In this paper, we update the expressions of that TVAR model to a fully multivariate setting (see Section \ref{sec:multivar_tvar}), meaning that the functional forms are defined via matrix notation for each element of the process vector. Next, we propose a method to improve the modeling capabilities of the original TVAR expressions, which allows them to characterize both the mean and the covariance of the original time-series data (see Section \ref{sub_sec_decoupling_trick}).

In the following subsections, the TVAR model of \cite{tvarnonstatspl} is revisited and the derivation of the updated multivariate expressions are presented. 

\subsection{Revisiting the TVAR Model and Deriving the Multivariate Expressions} \label{sec:multivar_tvar}

A TVAR model for multivariate time series can be described by an $M$-dimensional vector $\mathbf{x}(n)$ varying in time according to the following autoregressive expression:
\begin{equation}
\begin{array}{ccc}
\mathbf{x}(n+1) = \mathbf{A}(n)\mathbf{x}(n) + \mathbf{B}(n)\mathbf{v}(n)
\end{array}
\label{eq:tvar1_multivar}
\end{equation}
where $n$ is the time variable, $\mathbf{v}(n) \in \mathbb{R}^M$ is the perturbation noise, and $\mathbf{A}(n)$ and $\mathbf{B}(n)$ are diagonal matrices such that
\begin{equation}
\mathbf{A}(n) = \begin{bmatrix}
     a_1(n) &  \cdots & 0      \\
           \vdots &  \ddots & \vdots \\
                0 &  \cdots & a_M(n)
  \end{bmatrix}
\label{eq:tvar_a_diag_mat}
\end{equation}
and
\begin{equation}
\mathbf{B}(n) = \begin{bmatrix}
     b_1(n) &  \cdots & 0      \\
           \vdots &  \ddots & \vdots \\
                0 &  \cdots & b_M(n)
  \end{bmatrix}
\label{eq:tvar_b_diag_mat}.
\end{equation}
In Eqs. \eq{eq:tvar_a_diag_mat} and \eq{eq:tvar_b_diag_mat}, $a_{m}(n)$ and $b_{m}(n)$ are time-varying TVAR parameters describing the $m^{\text{th}}$ ``channel'' (i.e., $x_{m}(n)$) of the multivariate time series $\mathbf{x}(n)$ in \eq{eq:tvar1_multivar}. They are defined as
\begin{equation}
a_{m}(n) = \displaystyle\sum_{q=0}^{Q} \alpha_{m,q} f_{m,q}(n) 
\label{eq:tvar_basis_func_a}
\end{equation}
and
\begin{equation}
b_{m}(n) = \displaystyle\sum_{q=0}^{Q} \beta_{m,q} f_{m,q}(n) 
\label{eq:tvar_basis_func_b}
\end{equation}
with $f_{m,0}(n),...,f_{m,Q}(n)$ being scalar basis functions controlling the evolution of the parameters, and $\alpha_{m,0},...,\alpha_{m,Q}$ and $\beta_{m,0}, ..., \beta_{m,Q}$ being their weights.

The TVAR model described above should follow assumptions A3 and A4 as defined in \cite{daug_tvar_phm_2023}, namely:

A3) The initial value of the process $\mathbf{x}(0)$ is known and can be regarded as an arbitrary free parameter defined by the user. \\
A4) the noise $\mathbf{\text{v}}(n)$ is zero-mean, stationary, and has uncorrelated samples for $n\neq m$, i.e., $\expect\left[ \mathbf{\text{v}}(n)\mathbf{\text{v}}^{\text{T}}(m) \right]=\mathbf{0}$ if $n\neq m$.

Moreover, in this paper we consider that the following assumption should also hold: 

A5) The noise $\mathbf{\text{v}}(n)$ has a diagonal correlation matrix $\mathbf{\Phi}=\expect\left[ \mathbf{\text{v}}(n)\mathbf{\text{v}}^{\text{T}}(n) \right]$, i.e., $\expect\left[ v_i(n)v_j(n) \right]=0$ for $i \neq j$.

Considering the assumptions above and Eqs. \eq{eq:tvar_basis_func_a} and \eq{eq:tvar_basis_func_b}, it can be shown (refer to the Appendix section) that the mean vector and covariance matrix of $\mathbf{x}(n)$ in Eq. \eq{eq:tvar1_multivar} are given, respectively, as follows:
%
\begin{equation}
\mathbf{m}(n) = \displaystyle\prod_{k=0}^{n}\mathbf{A}(k) \mathbf{x}(0)
\label{eq:tvar_mean_multivar}
\end{equation}
and
\begin{equation}
\mathbf{C}(n) = \left\{ \mathbf{B}^2(n) + \sum_{l=0}^{n-1}\left[ \mathbf{B}(l)\prod_{k=l+1}^{n} \mathbf{A}(k) \right]^{2}  \right\} \mathbf{\Phi}.
\label{eq:tvar_cov_multivar}
\end{equation}

The same TVAR basis function and weights of \cite{daug_tvar_phm_2023} are employed here. However, now they should be defined for each diagonal element of $\mathbf{A}(n)$ and $\mathbf{B}(n)$. More precisely, let $Q=\alpha_{m,0}=\beta_{m,1}=1$ and $\alpha_{m,1}=\beta_{m,0}=0$ $\forall$ $m$ in Eqs. \eq{eq:tvar_basis_func_a} and \eq{eq:tvar_basis_func_b} and consider the scalar basis functions
\begin{equation}
f_{m,0}(n) = \frac{p_{m}(n+1)\text{e}^{r_{m,1}^{n+1}}}{p_{m}(n)\text{e}^{r_{m,1}^{n}}}
\label{eq:tvar_config1_f0_multivar}
\end{equation}
and
\begin{equation}
f_{m,1}(n) = \lambda_{m} p_{m}(n+1) \text{e}^{r_{m,1}^{n+1}} \sqrt{\frac{(1-r_{m,2}^{n+1})^2}{\text{e}^{2r_{m,1}^{n+1}}} - \frac{(1-r_{m,2}^{n})^2}{\text{e}^{2r_{m,1}^{n}}}}.
\label{eq:tvar_config1_f1_multivar}
\end{equation}
%
The basis functions above depend on the TVAR interpolation function $p_{m}(n)$. This is a functional form describing the time-varying behavior of the first- and second-order moments of $x_{m}(n)$, the $m^{\text{th}}$ component of $\mathbf{x}(n) \in \mathbb{R}^{M}$. This is one key difference from \cite{daug_tvar_phm_2023}, where a single scalar functional form $p(n)$ dictated the behavior over time of all channels of $\mathbf{x}(n)$. The user can choose $p_{m}(n)$ arbitrarily, as long as it fulfills the requirement R1) $p_m(n) \neq 0$ $\forall$ $m,n$. Furthemore, parameters $r_{m,1}$ and $r_{m,2}$ are rates of convergence and $\lambda_{m}$ is a constant gain. These parameters are also related to $x_{m}(n)$ and should fulfill requirements  R2) $0 < \{r_{m,1}, r_{m,2}\} < 1$ and R3) $\lambda_{m} \neq 0$. 

Having presented the chosen parameters and basis functions to customize the multivariate TVAR model, we proceed to introduce a set of time-varying matrices computed based on the defined parameters.
We call these the TVAR parameter matrices, which are necessary to derive the expressions employed in the next section. They are defined in Eqs. \eq{eq: P_matrix} to \eq{eq: Gamma_matrix}, where $p_{m}(n)$ is the $m^{\text{th}}$ TVAR interpolation function as defined above. Parameters $\lambda_m$, $r_{m,1}$ and $r_{m,2}$ are employed analogously.
\begin{equation}
\mathbf{P}(n) = \begin{bmatrix}
     p_1(n) &  \cdots & 0      \\
           \vdots &  \ddots & \vdots \\
                0 &  \cdots & p_M(n)
  \end{bmatrix},
\label{eq: P_matrix}
\end{equation}
\begin{equation}
\mathbf{\Lambda} = \begin{bmatrix}
     \lambda_{1} &  \cdots & 0      \\
           \vdots &  \ddots & \vdots \\
                0 &  \cdots & \lambda_{M}
  \end{bmatrix}
\label{eq: Lambda_matrix}
\end{equation}
\begin{equation}
\mathbf{R}(n) = \begin{bmatrix}
     1-r_{1,2}^{n} &  \cdots & 0      \\
           \vdots &  \ddots & \vdots \\
                0 &  \cdots & 1-r_{M,2}^{n}
  \end{bmatrix},
\label{eq: R2_matrix}
\end{equation}
\begin{equation}
\mathbf{S}(n) = \begin{bmatrix}
     \text{e}^{r_{1,1}^{n}} &  \cdots & 0      \\
           \vdots &  \ddots & \vdots \\
                0 &  \cdots & \text{e}^{r_{M,1}^{n}}
  \end{bmatrix}.
\label{eq: R1_matrix}
\end{equation}
\begin{equation}
\mathbf{\Gamma} = \begin{bmatrix}
     \displaystyle\frac{\text{e}^{-1}}{p_{1}(0)} &  \cdots & 0      \\
           \vdots &  \ddots & \vdots \\
                0 &  \cdots & \displaystyle\frac{\text{e}^{-1}}{p_{\text{M}}(0)}
  \end{bmatrix},
\label{eq: Gamma_matrix}
\end{equation}

It can be shown (refer to the Appendix section) that the general-term formula of Eq. \eq{eq:tvar1_multivar} and its first- and second-order moments can be written as functions of Eqs. \eq{eq: P_matrix} to \eq{eq: Gamma_matrix} when employing the aforementioned TVAR basis functions and weights. More specifically, by substituting Eqs. \eq{eq:tvar_config1_f0_multivar} and \eq{eq:tvar_config1_f1_multivar} into Eqs. \eq{eq:tvar_basis_func_a} and \eq{eq:tvar_basis_func_b}, using the resulting expressions for $\mathbf{A}(n)$ and $\mathbf{B}(n)$ in Eq. \eq{eq:tvar1_multivar}, and rearranging the process expression, we obtain the following equation:
\begin{equation}
\small
\begin{array}{l}
\mathbf{x}(n+1) =  \mathbf{P}(n+1)\mathbf{S}(n+1) \Big\{ \tilde{\mathbf{x}}(0) +  \\
 \mathbf{\Lambda}\displaystyle\sum_{l=0}^{n}\left\{ [\mathbf{R}(l+1)\mathbf{S}^{-1}(l+1)]^2 - [\mathbf{R}(l)\mathbf{S}^{-1}(l)]^2 \right\}^{\frac{1}{2}}\mathbf{v}(l) \Big\}
\end{array}
\label{eq:tvar_general_term_form_updated}
\end{equation}
where it has been assumed that:
\begin{equation}
\tilde{\mathbf{x}}(0)=\mathbf{\Gamma}\mathbf{x}(0)
\label{eq:x_scaled_by_gamma}
\end{equation}

In Eq. \eq{eq:x_scaled_by_gamma}, $\tilde{\mathbf{x}}(0)$ is the scaled (by matrix $\mathbf{\Gamma}$) version of vector $\mathbf{x}(0)$ (refer to assumption A3). The mean vector and covariance matrix of \eq{eq:tvar_general_term_form_updated} are given respectivelly by:
\begin{equation}
\mathbf{m}(n) = \mathbf{P}(n) \mathbf{S}(n) \tilde{\mathbf{x}}(0)
\label{eq:tvar_mean_multivar_updated}
\end{equation}
and  
\begin{equation}
\mathbf{C}(n) = \mathbf{P}^2(n)\mathbf{R}^2(n) \mathbf{\Lambda}^2 \mathbf{\Phi}.
\label{eq:tvar_cov_multivar_updated}
\end{equation}
%

As $n$ grows, it can be verified that  $\mathbf{S}(n)$ and $\mathbf{R}(n)$ converge to the identity matrix at speeds controlled by parameters $r_{m,1}$ and $r_{m,2}$. This makes Eqs. \eq{eq:tvar_mean_multivar_updated} and \eq{eq:tvar_cov_multivar_updated} reduce to their steady-state formulas characterized by 
\begin{equation}
\mathbf{m}_{\text{ss}}(n) = \mathbf{P}(n) \tilde{\mathbf{x}}(0)
\label{eq:tvar_mean_multivar_ss_updated}
\end{equation}
and
\begin{equation}
\mathbf{C}_{\text{ss}}(n) = \mathbf{P}^2(n) \mathbf{\Lambda}^2\mathbf{\Phi}.
\label{eq:tvar_cov_multivar_ss_updated}
\end{equation}

It can be noted that Eqs. \eq{eq:tvar_mean_multivar_updated} and \eq{eq:tvar_cov_multivar_updated} consist of the product of three terms, a gain (i.e., $\tilde{\mathbf{x}}(0)$ and $\mathbf{\Lambda}^2$), a convergence matrix (i.e., $\mathbf{S}(n)$ and $\mathbf{R}(n)$) and a functional form $\mathbf{P}(n)$ characterizing the steady-state behavior of both equations.

According to Eqs. \eq{eq:tvar_mean_multivar_updated} and \eq{eq:tvar_cov_multivar_updated}, the multivariate TVAR process cannot model non-stationary multivariate 
time series whose mean and covariance vary indepedently of each other over time, as their stochastic expressions are both dependent on $\mathbf{P}(n)$.
 The same issue was observed in \cite{daug_tvar_phm_2023} when deriving the univariate time series stochastic expressions.
To circumvent this problem, in \cite{daug_tvar_phm_2023}, it was proposed to use two TVAR sub-process expressions, one to model the mean and other the covariance of time-series data to be augmented. 
These TVAR expressions were employed to generate synthetic samples for the data augmentation process that captured the dynamics of the first- and second-order non-stationarities in the data separately. 

In this work, we propose a method to obtain multivariate TVAR process expressions like Eq. \eq{eq:tvar_general_term_form_updated} such that the mean and covariance for such process can be defined independently from each other. We name this method the TVAR decoupling trick, which is explained in the following subsection.

\subsection{The TVAR Decoupling Trick} \label{sub_sec_decoupling_trick}

Here we show how Eq. \eq{eq:tvar_general_term_form_updated} can be modified so that the expressions for its mean and covariance are not functions of $\mathbf{P}(n)$. We start by borrowing the idea of \cite{daug_tvar_phm_2023} of defining two TVAR sub-processes to represent respectively the first- and second-order moments of the underlying stochastic process. These two TVAR sub-processes are named here $\mathbf{x}_{1}(n)$ and $\mathbf{x}_{2}(n)$ and are computed as in \eq{eq:tvar_general_term_form_updated}.

The sub-processes $\mathbf{x}_{1}(n)$ and $\mathbf{x}_{2}(n)$ have their own set of TVAR parameter matrices $\mathbf{P}_{i}(n),\mathbf{S}_{i}(n),\mathbf{R}_{i}(n),\mathbf{\Gamma}_{i}(n),\mathbf{\Lambda}_{i}(n)$ for $i=1,2$. Moreover, $\mathbf{x}_{1}(n)$ and $\mathbf{x}_{2}(n)$ have their own expressions for the mean ($\mathbf{m}_{1}(n+1)$ and $\mathbf{m}_{2}(n+1)$) and covariance ($\mathbf{C}_{1}(n+1)$ and $\mathbf{C}_{2}(n+1)$) that follow Eqs. \eq{eq:tvar_mean_multivar_updated} and \eq{eq:tvar_cov_multivar_updated}, respectively.

Let us define the composed process
\begin{equation}
\mathbf{x}'(n) = \mathbf{x}_{1}(n) + \mathbf{x}_{2}(n)
\label{eq:tvar_composed_process}
\end{equation}
with mean $\mathbf{m}'(n)$ and covariance $\mathbf{C}'(n)$. Due to the linearity of the expectation operator, it is straightforward to show that:
\begin{equation}
\mathbf{m}'(n) = \mathbf{m}_{1}(n) + \mathbf{m}_{2}(n).
\label{eq:tvar_mean_multivar_composed}
\end{equation}
Moreover, based on assumption A4), it can be shown that:
\begin{equation}
\mathbf{C}'(n) = \mathbf{C}_{1}(n) + \mathbf{C}_{2}(n).
\label{eq:tvar_cov_multivar_composed}
\end{equation}
The idea behind the decoupling trick is to choose TVAR parameter matrices that make $\mathbf{m}_{2}(n)=\mathbf{0}$ in Eq. \eq{eq:tvar_mean_multivar_composed} and $\mathbf{C}_{1}(n)=\mathbf{0}$ in \eq{eq:tvar_cov_multivar_composed}, which result in $\mathbf{m}'(n) = \mathbf{m}_1(n)$ and $\mathbf{C}'(n) = \mathbf{C}_2(n)$. By doing so, we can ensure that the theoretical two first moments of $\mathbf{x}'(n)$ are represented by two separate sets of TVAR parameter matrices. To show how that can be achieved, we rewrite Eqs. \eq{eq:tvar_mean_multivar_composed} and \eq{eq:tvar_cov_multivar_composed} as follows:
\begin{equation}
\mathbf{m}'(n) = \mathbf{P}_{1}(n) \mathbf{S}_{1}(n) \tilde{\mathbf{x}}_{1}(0) + \mathbf{P}_{2}(n) \mathbf{S}_{2}(n) \tilde{\mathbf{x}}_{2}(0)
\label{eq:tvar_mean_multivar_composed_updated}
\end{equation}
and
\begin{equation}
\mathbf{C}'(n) = \mathbf{P}_{1}^2(n)\mathbf{R}_{1}^2(n) \mathbf{\Lambda}_{1}^2 + \mathbf{P}_{2}^2(n)\mathbf{R}_{2}^2(n) \mathbf{\Lambda}_{2}^2.
\label{eq:tvar_cov_multivar_composed_updated}
\end{equation}

In Eq. \eq{eq:tvar_mean_multivar_composed_updated}, setting $\tilde{\mathbf{x}}_{2}(0) = \mathbf{0}$ would result in $\mathbf{m}'(n) = \mathbf{m}_1(n)$. This would be equivalent to setting $\mathbf{x}_{2}(0)=0$ (see Eq. \eq{eq:x_scaled_by_gamma}). In turn, $\mathbf{C}'(n) = \mathbf{C}_2(n)$ could be achieved by making $\mathbf{\Lambda}_{1} = \mathbf{0}$ in Eq. \eq{eq:tvar_cov_multivar_composed_updated}. However, this last equality is not possible given the requirement set to the $\lambda_m$ constants that make up the TVAR parameter matrix $\mathbf{\Lambda}$ (refer to requirement R3 and Eq. \eq{eq: Lambda_matrix}). To circumvent this problem, we propose to choose $\mathbf{\Lambda}_{1} \approx \mathbf{0}$, but not necessarily equal to the zero vector, in a way that $\mathbf{\Lambda}_{2} \gg \mathbf{\Lambda}_{1}$ and
\begin{equation}
\mathbf{P}_{1}^2(n)\mathbf{R}_{1}^2(n) \mathbf{\Lambda}_{1}^2 \ll \mathbf{P}_{2}^2(n)\mathbf{R}_{2}^2(n) \mathbf{\Lambda}_{2}^2
\label{eq:tvar_cov_innequality}
\end{equation}
As the terms in \eq{eq:tvar_cov_innequality} model a covariance matrix, all of them are positive. Thus, picking very small values for $\mathbf{\Lambda}_{1}$ would suffice to ensure \eq{eq:tvar_cov_innequality} holds. Assuming this approximation to be true, and that the matrices can be defined so that convergence to the identity matrix happens sufficiently fast, i.e. assuming:
\begin{equation}
\mathbf{S}_{2}(n), \mathbf{R}_{1}(n)\approx \mathbf{I}
\label{eq:identity_for_small_n}
\end{equation}
holds for small $n$, the general-term expression of Eq. \eq{eq:tvar_composed_process} can be written as follows:
\begin{equation}
\small
\begin{array}{l}
\mathbf{x}(n+1) =  \mathbf{S}_1(n+1)\mathbf{P}_1(n+1)\tilde{\mathbf{x}}(0) +  \\
 \mathbf{\Lambda}_2\displaystyle\sum_{l=0}^{n}\left\{ [\mathbf{R}_2(l+1)\mathbf{S}_2^{-1}(l+1)]^2 - [\mathbf{R}_2(l)\mathbf{S}_2^{-1}(l)]^2 \right\}^{\frac{1}{2}}\mathbf{v}(l) 
\end{array}
\label{eq:tvar_general_term_form_updated2}
\end{equation}
with mean vector and covariance matrix corresponding respectively to:
\begin{equation}
\mathbf{m}(n) = \mathbf{P}_{1}(n) \mathbf{S}_{1}(n) \tilde{\mathbf{x}}_{1}(0)
\label{eq:tvar_mean_multivar_composed_updated_final}
\end{equation}
and
\begin{equation}
\mathbf{C}(n) = \mathbf{P}_{2}^2(n)\mathbf{R}_{2}^2(n) \mathbf{\Lambda}_{2}^2.
\label{eq:tvar_cov_multivar_composed_updated_final}
\end{equation}
The actual values of the TVAR parameter matrices used to achieve those approximations are detailed in the next section. 

In Eqs. \eq{eq:tvar_general_term_form_updated2} to \eq{eq:tvar_cov_multivar_composed_updated_final}, the notation was simplified by removing the prime symbol ($'$) when referring to the new process and corresponding mean vector and covariance matrix as defined respectively in Eq. \eq{eq:tvar_composed_process} to \eq{eq:tvar_cov_multivar_composed}. From this point until the end of paper, $\mathbf{x}(n)$, $\mathbf{m}(n)$, and $\mathbf{C}(n)$ correspond respectively to the definitions in Eqs. \eq{eq:tvar_general_term_form_updated2}, \eq{eq:tvar_mean_multivar_composed_updated_final}, and \eq{eq:tvar_cov_multivar_composed_updated_final}. 

Similarly to Eqs. \eq{eq:tvar_mean_multivar_ss_updated} and \eq{eq:tvar_cov_multivar_ss_updated}, it can be shown that, in steady state, Eqs. \eq{eq:tvar_mean_multivar_composed_updated_final} and \eq{eq:tvar_cov_multivar_composed_updated_final} reduce to
\begin{equation}
\mathbf{m}_{\text{ss}}(n+1) = \mathbf{P}_{1}(n) \tilde{\mathbf{x}}_{1}(0)
\label{eq:tvar_mean_multivar_composed_updated_ss}
\end{equation}
and
\begin{equation}
\mathbf{C}_{\text{ss}}(n+1) = \mathbf{P}_{2}^2(n) \mathbf{\Lambda}_{2}^2,
\label{eq:tvar_cov_multivar_composed_updated_ss}
\end{equation}
which can be defined independently from each other.

Moreover, it can be seen that, in Eq. \eq{eq:tvar_general_term_form_updated2}, parameters $\tilde{\mathbf{x}}_{1}(0)$, $\tilde{\mathbf{S}}_{1}(n)$ and $\mathbf{P}_1(n)$, control the gain, convergence and steady-state functional form of the theoretical mean in Eq. \eq{eq:tvar_mean_multivar_composed_updated_final}, which only depends on these parameters. Similarly, $\mathbf{\Lambda}_2$, $\mathbf{R}_2(n)$, and $\mathbf{P}_2(n)$ determine solely the behavior of the covariance expression in Eq. \eq{eq:tvar_cov_multivar_composed_updated_final}. Thus, by using the proposed approach, we can now define TVAR processes with mean vectors which are decoupled from the covariance matrices.

Having presented how the new time series model can be obtained by means of the proposed TVAR decoupling trick, we detail ahead how the data augmentation is achieved based on this model, as well as the procedure to obtain the TVAR parameter matrices $\mathbf{P}_{1}(n)$ and $\mathbf{P}_{2}(n)$.

\section{Data Augmentation with the TVAR Model} \label{sec:data_augmentation}

\subsection{Overview of the Proposed Method}

The procedure for achieving data augmentation is similar to the one described in \cite{daug_tvar_phm_2023}, but with some differences. More precisely, here the new TVAR model of Eq. \eq{eq:tvar_general_term_form_updated2} is employed to represent the underlying stochastic process of the multivariate time series of interest. We assume that this stochastic process can be characterized by empirical statistics computed from the time series data. Fitting the new TVAR model amounts to finding TVAR parameter matrices that make the theoretical first- and second-order moment expressions of Eqs. \eq{eq:tvar_mean_multivar_composed_updated_final} and \eq{eq:tvar_cov_multivar_composed_updated_final} match the empirical statistic expressions calculated for the data. Those parameter matrices are then employed in the TVAR process equation (Eq. \eq{eq:tvar_general_term_form_updated2}) to create synthetic data samples.

\subsection{Setting up the TVAR Parameter Matrices} \label{sec:interp_functions}

The procedure to determine TVAR parameter matrices can be divided into two steps. First, the gain and convergence parameters are defined in a way that ensures the TVAR decoupling behavior described in Section \ref{sub_sec_decoupling_trick}. Next, the matrices $\mathbf{P}_{1}(n)$ and $\mathbf{P}_{2}(n)$ with the TVAR interpolation functions characterizing the behavior of the mean and covariance in steady state (Eqs. \eq{eq:tvar_mean_multivar_composed_updated_ss} and \eq{eq:tvar_cov_multivar_composed_updated_ss}) are calculated from the empirical statistics computed from the data. Ahead, we address the choice of the gain and convergence parameters.

\subsubsection{Choosing the Gain and Convergence Parameters} \label{sec:gain_and_convergence}

As discussed in Section \ref{sub_sec_decoupling_trick}, the parameters of the TVAR sub-processes $\mathbf{x}_{1}(n)$ and $\mathbf{x}_{2}(n)$ should be chosen to ensure $\mathbf{m}_{2}(n) = \mathbf{0}$ and $\mathbf{C}_{1}(n) \approx \mathbf{0}$. To achieve the former, we make $\tilde{\mathbf{x}}_{2}(0)=\mathbf{0}$. To approximate the latter, we pick small values of $r_{m,1}$ to build $\mathbf{\Lambda}_1$ (see Eqs. \eq{eq: Lambda_matrix} and \eq{eq:tvar_cov_multivar_composed_updated}), e.g. $r_{m,1} = 0.01$ for $m=1,...,M$. Furthermore, to make Eq. \eq{eq:identity_for_small_n} valid for small $n$, we built the parameter matrix  $\mathbf{S}_{2}(n)$ (see Eq. \eq{eq: R1_matrix}) by choosing $r_{m,2} = 0.01$ for $m=1,...,M$. 

The remaining quantities to be defined are the gain terms of the mean and covariance (see Eqs. \eq{eq:tvar_mean_multivar_composed_updated_final} to \eq{eq:tvar_cov_multivar_composed_updated_ss}), and the process noise $\mathbf{v}(n)$. In this work, for the sake of simplicity, we choose $\tilde{\mathbf{x}}_{1}(0) = \mathbf{1}$ and $\mathbf{\Lambda}_{2}(n) = \mathbf{I}$, which make Eqs.\eq{eq:tvar_mean_multivar_composed_updated_ss} and \eq{eq:tvar_cov_multivar_composed_updated_ss} reduce to $\mathbf{m}_{\text{ss}}(n) = \mathbf{P}_{1}(n)$ and $\mathbf{C}_{\text{ss}}(n) = \mathbf{P}_{2}^{2}(n)$. The noise $\mathbf{v}(n)$ is defined by its correlation matrix $\mathbf{\Phi}$ (assumptions A4 and A5) and has no \textit{a priori} constraints regarding its probability distribution. Here, we define $\mathbf{v}(n)$ as a white Gaussian noise (WGN) with zero mean and standard deviation equal to 0.1 for all the $m=1,...,M$ elements. Next, we discuss how $\mathbf{P}_{1}(n)$ and $\mathbf{P}_{2}(n)$ can be obtained.

\subsubsection{Computing $\mathbf{P}_{1}(n)$ and $\mathbf{P}_{2}(n)$} \label{sec:empirical_stats}

The parameter matrices $\mathbf{P}_{1}(n)$ and $\mathbf{P}_{2}(n)$ are made equal to the empirical first- and second-order statistics computed from the data. Because the multivariate time series are considered to be equally-sampled (assumption A1), the empirical mean and covariance of the time series data are computed through ensemble averaging. Given a dataset of $M$-dimensional multivariate time series observed for $n=1,...,N$ points in time, the $j^{\text{th}}$ time series of this dataset can be defined as
\begin{equation}
\mathbf{Y}_j = [\mathbf{y}_j(0),...,\mathbf{y}_j(N-1)]^{\text{T}} \text{ with } \mathbf{y}_j(n) \in \mathbb{R}^M
\label{multivariate_time_series}
\end{equation}
and the time series dataset 
\begin{equation}
\mathcal{D} = \{\mathbf{Y}_j\}_{j=1,...,J}
\label{collection_sigs}
\end{equation}
Based on $\mathcal{D}$, the empirical mean vector and covariance matrix at time $n$ can be calculated via ensemble averaging as
\begin{equation}
\overline{\mathbf{m}}(n)=\frac{1}{J}\sum_{j=1}^{J} \mathbf{y}_j(n)
\label{emp_mean_vector}
\end{equation}
and
\begin{equation}
\overline{\mathbf{C}}(n) = 
\displaystyle\frac{1}{J}\sum_{j=1}^{J} 
\left[\mathbf{y}_{j}(n) - \overline{\mathbf{m}}(n)\right]\left[\mathbf{y}_{j}(n) - \overline{\mathbf{m}}(n)\right]^{\text{T}}
\label{emp_cov_matrix}
\end{equation}
respectively. Based on Eqs. \eq{emp_mean_vector} and \eq{emp_cov_matrix}, the $\mathbf{P}_{1}(n)$ and $\mathbf{P}_{2}(n)$ matrices can be computed as follows (Eq. \eq{eq: P_matrix}):
\begin{equation}
\mathbf{P}_1(n) = \begin{bmatrix}
     \overline{m}_1(n) &  \cdots & 0      \\
           \vdots &  \ddots & \vdots \\
                0 &  \cdots & \overline{m}_M(n)
  \end{bmatrix}
\label{eq: P1_matrix_fitted}
\end{equation}
and
\begin{equation}
\mathbf{P}_2(n) = \begin{bmatrix}
     \overline{C}_{1,1}(n) &  \cdots & 0      \\
           \vdots &  \ddots & \vdots \\
                0 &  \cdots & \overline{C}_{M,M}(n)
  \end{bmatrix}
\label{eq: P2_matrix_fitted}
\end{equation}
where $\overline{m}_1(n),...,\overline{m}_M(n)$ and $\overline{C}_{1,1}(n),...,\overline{C}_{M,M}(n)$ are elements of $\overline{\mathbf{m}}(n)$ and $\overline{\mathbf{C}}(n)$, respectively.

\subsubsection{Data Augmentation Procedure} \label{sec:daug_proc}

Algorithm \ref{ref:my_algo} corresponds to the proposed data augmentation method. This algorithm is simpler than the one proposed in \cite{daug_tvar_phm_2023}, as the modeling of the multivariate time series is performed by a single multivariate TVAR model. The following section presents the experiments carried out to empirically evaluate the new time series model and the data augmentation algorithm.
\begin{algorithm}[!htb]

\SetAlgoLined

\hrule 
\hrule
\hrule

\BlankLine

\KwIn{Dataset $\mathcal{D}=\{\mathbf{Y}_j\}_{j=1,...,J}$ of multivariate time series to augment (see Eq. \eq{collection_sigs})}

\BlankLine

\KwOut{Dataset $\mathcal{D}_{\text{aug}}$ of augmented time series.}

\BlankLine

\hrule 
\hrule
\hrule

\BlankLine

\textbf{Step 1:} Initialize TVAR parameter matrices initial vectors as detailed in Section \ref{sec:gain_and_convergence}.

\BlankLine

Initialize the perturbation noise correlation matrix $\mathbf{\Phi}$ as detailed in Section \ref{sec:gain_and_convergence}, and following A4 and A5.

\BlankLine

Initialize time series length $N$ and number of augmented samples to create $L$;

\BlankLine

\textbf{Step 2:} For $\mathcal{D}$, compute empirical statistics $\overline{\mathbf{m}}(n)$ and $\overline{\mathbf{C}}(n)$ using Eqs. \eq{emp_mean_vector} and \eq{emp_cov_matrix}\;

\BlankLine

Store $\overline{\mathbf{m}}(n)$ and $\overline{\mathbf{C}}(n)$\;

\BlankLine

\textbf{Step 3:} Use $\overline{\mathbf{m}}(n)$ and $\overline{\mathbf{C}}(n)$ to compute $\mathbf{P}_1(n)$ and $\mathbf{P}_2(n)$ (See Eqs. \eq{eq: P1_matrix_fitted} and \eq{eq: P2_matrix_fitted})\;

\BlankLine

Store $\mathbf{P}_1(n)$ and $\mathbf{P}_2(n)$\;

\BlankLine

\textbf{Step 4:} Use $\mathbf{\Gamma}$ to compute the vector $\tilde{\mathbf{x}}_1(0)=\mathbf{\Gamma}\mathbf{x}_1(0)$ (see Eq. \eq{eq: Gamma_matrix}) \;

\BlankLine

\textbf{Step 5:} Create the TVAR function for $\mathbf{x}(n)$ by using the quantites computed in the previous steps and Eq. \eq{eq:tvar_general_term_form_updated}\;

\BlankLine

\textbf{Step 6:} Nested loop

\BlankLine

\For{$l=1$ \KwTo $L$}{

    \Indp

    \For{$n=0$ \KwTo $N-1$}{
        Compute $\mathbf{x}(n)$ \;
    }
    \Indm
    
    Append $\mathbf{x}(n)$ to $\mathcal{D}_{\text{aug}}$\;
}

\BlankLine

\textbf{Return:} $\mathcal{D}_{\text{aug}}$ \;

\BlankLine

\hrule 
\hrule
\hrule

\BlankLine

\caption{Data augmentation method using the proposed multivariate TVAR model.}
\label{ref:my_algo}
\end{algorithm}

\section{Experimental Study} \label{sec:exp_study}

This section presents the empirical evaluation of the proposed data augmentation method when applied to improve the performance of failure prognostics solutions. The C-MAPSS datasets \cite{saxena2008turbofan}, extensively employed for testing failure prognostics solutions \cite{ramasso2014review}, have been used for this evaluation. Datasets \#1 (FD001) and \#3 (FD003) were chosen for the tasks for the sake of simplicity, as they correspond to a single operating condition. Adjustments have also been made to the number of samples so that the simulated data could more closely resemble the data scarcity problems associated with real-world failure prognostics. Namely, it was assumed that only five samples were available in each experiment. Therefore, for each run of the experiments described below, five samples, out of the 100 available, have been randomly drawn from the corresponding dataset.

\subsection{AutoML for Prognostics} \label{subsec:exp_automl}

The AutoML approach for definition of failure prognostics methods proposed in \cite{kefalas2021automated} is employed here. It was chosen as it consists of a fully automated pipeline to design, train and test ML models for prognostics, being therefore a good choice for the assessment of data-centric approaches. The method was applied as proposed in the original paper, with the implementation of the data processing pipeline developed based on the Python code that accompanies it. A summary of the methodology is described here but the reader should refer to the original paper for more details. It consists of the following steps:
\begin{enumerate}
    \item \textbf{Pre-processing}: pre-processing steps comprise sensor selection and normalization. Concerning the former, for the datasets employed here (\#1 and \#3) 14 out of the 21 available sensors, namely sensors \#2, 3, 4, 7, 8, 9, 11, 12, 13, 14, 15, 17, 20 and 21, are employed. This is similarly performed in other related works using the same datasets. In terms of normalization, all data is transformed to zero mean and unit variance. 
    \item \textbf{RUL data preparation}: the points in time at which RUL estimation is evaluated are defined based on a fixed time window increment. A time window increment of 10 was used for the experiments. At each of those points, the time window of sensor data from the beginning of operation up to that point in time is associated with a ground truth RUL value for training/testing of models. The strategy of defining the RUL values over time in a piecewise linear fashion is employed. This means the RUL is fixed to a constant value during the beginning of the equipment's life and only after a predefined point in time it begins to linearly reduce towards zero. Constant RUL values for the datasets \#1 and \#3 employed in this study are respectively 115 and 125.  
    \item \textbf{Feature extraction and selection}: The whole time window of data available until the point in time when the RUL is evaluated is employed for feature extraction. This means a multivariate time window of sensor data of arbitrary length is transformed into a feature vector of constant size. Features are defined based on the Python package \textit{tsfresh} \cite{tsfreshpaper}
    \item \textbf{RUL estimation model training/testing}: after execution of the previous steps, the RUL estimation is transformed into a standard regression problem where features are inputs and RUL is the output. This is the step where AutoML is employed for automating both the ML algorithm selection and hyperparameter optimization. The Python package \textit{TPOT} is used for this purpose \cite{le2020scaling}. The package employs genetic programming to perform the AutoML task. Although this approach provides a good framework for testing data-centric methods due to the automation of the complete process for designing the ML model, it must be noted that the AutoML process can be computationally expensive, limiting the number of experiments that can be performed for a certain computation budget.
\end{enumerate}

Metrics employed for assessing the prognostics solution performance are also the same employed in \cite{kefalas2021automated}, consisting of RMSE and the scoring function used in the PHM Society data challenge associated with the initial release of the dataset. This scoring function ($S$) is presented in Eq. \eq{eq:scoring_function}. In the equation, $d_j = \hat{\text{RUL}}_j - \text{RUL}_j$, where $\hat{\text{RUL}}_j$ is the estimated value produced by the prognostics solution and $j$ is the sample number. It can be noticed that such metric penalizes more heavily estimates which are late than those which are early.
\begin{equation}
    S = 
    \begin{cases}
        \displaystyle\sum_{j=1}^{J}\text{e}^{(-d_j/13)}-1\; \text{if}\; d_j < 0\\
        \displaystyle\sum_{j=1}^{J}\text{e}^{(d_j/10)}-1\; \text{if}\; d_j \geq 0\\
    \end{cases}
    \label{eq:scoring_function}
\end{equation}

\subsection{Data augmentation Procedures} \label{subsec:exp_daug_procedures}

In order to apply the proposed data augmentation approach, TVAR parameters are obtained based on the empirical mean vector (Eq. \eq{emp_mean_vector}) and covariance matrix (Eq. \eq{emp_cov_matrix}) calculated from the historical data. For such calculations to be performed, the corresponding samples must be aligned according to time $n$ as presented in the equations. For application to failure prognostics, available historical sensor data have been aligned based on their associated RUL values, so $n$ in the TVAR equations corresponds to the RUL in this case. Following such definition, data augmentation can be performed as described in section \ref{sec:data_augmentation} and the resulting dataset can be employed for prognostics as described in section \ref{subsec:exp_automl}

\begin{figure*} [!htb]
\centering
\includegraphics[width=5in]{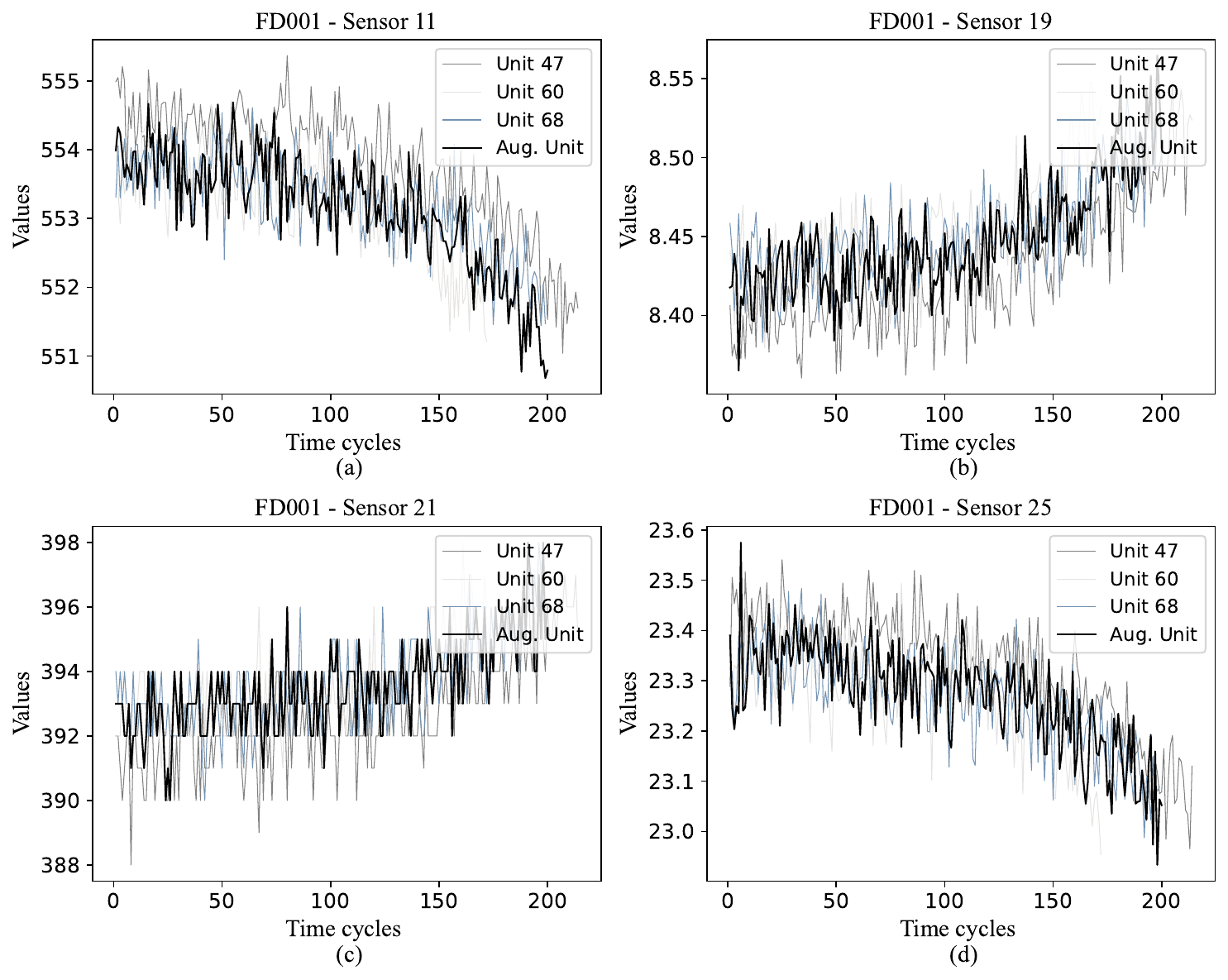}
\caption{Run to failure data from four sensors (11, 29, 21, and 25) for dataset \#1 (FD001). Each plot presents data from 3 real units (47, 60, and 68) and one time series resulting from data augmentation (Aug. Unit).}
\label{fig:interp_example}
\end{figure*}

Figure \ref{fig:interp_example} presents sample time series from four sensor for three units in the dataset \#1 (47, 60 and 68) and one sample corresponding to data augmentation (Aug. Unit). Time series resulting from the proposed data augmentation procedure cannot be distinguished from the real data based on visual inspection of the plots.

\subsection{Experiments and Results}
As previously mentioned, experiments were performed with the assumption that historical data from only five units was available, as this level of data scarcity is representative of many real world use cases. Two sets of experiments were performed, employing respectively datasets \#1 and \#3. Each set comprised 100 experiments and in each of such experiments five units were randomly selected from the dataset to serve as the data employed for augmentation and ML model training. In each experiment, a baseline was created using only the data from the five units to train the RUL estimation model. A second model was then trained with additional five samples generated using the proposed data augmentation method.

Performance for each experiment was assessed based on the RMSE and $S$ metrics described above. Table \ref{tab:results} presents the average relative improvement obtained as a result of the data augmentation for each set of experiments compared to the baseline with no augmentation. It can be noticed from the table that the proposed method resulted in a considerable performance improvement, especially for dataset \#3.

\begin{table}[]
    \centering
    \begin{tabular}{|c|c|c|}
        \hline
         Dataset & \makecell{RMSE avg. \\ improvement} & \makecell{$S$ avg. \\ improvement} \\
         \hline
         \#1 (FD001) & 2\% & 4\% \\
         \hline
         \#2 (FD003) & 6\% & 20\% \\
         \hline
    \end{tabular}
    \caption{Average relative improvement obtained in each set of experiments resulting from the proposed data augmentation approach when compared to the baseline with no augmentation.}
    \label{tab:results}
\end{table}

\section{Conclusion} \label{sec:conclusion}

This paper presented a novel method for non-stationary multivariate time series data augmentation and its application to failure prognostics. The method extends previous work from the authors \cite{daug_tvar_phm_2023} to employ time-varying autoregressive (TVAR) models to generate synthetic data based on mean and covariance estimates obtained from the available real samples. The method was successfully tested on a failure prognostics application. The C-MAPSS dataset, commonly employed for testing and benchmarking failure prognostics methods was used for the experiments. The AutoML approach originally proposed in \cite{kefalas2021automated} was applied here as the means for designing the prognostics methods. It was chosen as it automates the ML method selection and hyperparameter tuning, providing a practical and unbiased framework for testing data-centric prognostics solutions. The only drawback of such approach is its computational cost, as it results in a large search space which is explored based on genetic programming. Experiments comparing the results with and without data augmentation provided empirical evidence of the successful application of the proposed approach to improve failure prognostics performance under data scarcity situations.

The experiments were limited to a single scenario in terms of number of real and synthetic samples. Future work will include a larger variety of experiment conditions, as well as additional datasets, including C-MAPSS dataset \#2 and \#4 which correspond to varying operating conditions, and other more realistic datasets such as the one described in \cite{ariaschao2021newcmapss}. Future work will also include benchmarking with other data augmentation approaches for failure prognostics and experimenting with additional use cases, with focus on industrial applications using real-world data.

\bibliographystyle{apacite}
\bibliography{ijphm}

\section*{Appendix. Deriving Mathematical Expressions} \label{sec:appendixI}

\subsection*{Determining Eq. \eq{eq:tvar_mean_multivar}}

By repeated substitution, one can derive the general-term expression of $\mathbf{x}(n)$ from Eq. \eq{eq:tvar1_multivar} as follows:
\begin{displaymath}
\begin{array}{c}
\mathbf{x}(1) = \mathbf{A}(0)\mathbf{x}(0) + \mathbf{B}(0)\mathbf{v}(0) \\ \\
\mathbf{x}(2) = \mathbf{A}(1)\mathbf{x}(1) + \mathbf{B}(1)\mathbf{v}(1) \\
\mathbf{x}(2) = \mathbf{A}(1)\mathbf{A}(0)\mathbf{x}(0) + \mathbf{A}(1)\mathbf{B}(0)\mathbf{v}(0) + \mathbf{B}(1)\mathbf{v}(1) \\
\cdots
\end{array}
\end{displaymath}
\begin{equation}
\begin{array}{l}
\mathbf{x}(n+1) = \\
\displaystyle\mathbf{x}(0)\prod_{k=0}^{n}\mathbf{A}(k) +
\displaystyle \sum_{l=0}^{n-1} \mathbf{B}(l)\mathbf{v}(l)\prod_{k=l+1}^{n}\mathbf{A}(k) + \mathbf{B}(n)\mathbf{v}(n) 
\end{array}   
\label{eq:general_term_demonstration}
\end{equation}
where the above expression is valid for $n>0$. The mean vector of the TVAR process can be obtained by taking the expected value of Eq. \eq{eq:general_term_demonstration}, which results in
\begin{equation}
\begin{array}{l}
\mathbf{m}(n) = \displaystyle\mathbf{x}(0)\prod_{k=0}^{n}\mathbf{A}(k) + \\
\displaystyle \sum_{l=0}^{n-1} \mathbf{B}(l)\expect\{\mathbf{v}(l)\}\prod_{k=l+1}^{n}\mathbf{A}(k) + \mathbf{B}(n)\expect\{\mathbf{v}(n)\} 
\end{array}   
\label{eq:mean_from_gen_term}
\end{equation}
because $\mathbf{x}(0)$, $\mathbf{A}(n)$, and $\mathbf{B}(n)$ are deterministic (see Eqs. \eq{eq:tvar_a_diag_mat} and \eq{eq:tvar_b_diag_mat} and assumption A3). Since the noise vector is zero-mean (see assumption A4), every $\expect \left[ \mathbf{v}(l) \right]$ term should be zero in Eq. \eq{eq:mean_from_gen_term}, making this equation reduce to Eq. \eq{eq:tvar_mean_multivar}.
\subsection*{Determining Eq. \eq{eq:tvar_cov_multivar}}

The covariance matrix of the TVAR process can be calculated by subtracting Eq. \eq{eq:tvar_mean_multivar} from both sides of Eq. \eq{eq:general_term_demonstration}, taking the outer product, and computing its expected value. Then, by rearranging the resulting expression, we get
\begin{equation}
\small
\begin{array}{l}
\mathbf{C}(n) =
\expect\left\{\left[ \mathbf{B}(n)\mathbf{v}(n) \right]\left[ \mathbf{B}(n)\mathbf{v}(n) \right]^{\text{T}}\right\} + \\
\expect\left\{ \left[ \mathbf{B}(n)\mathbf{v}(n) \right]\left[ \displaystyle \sum_{l=0}^{n-1} \mathbf{B}(l)\mathbf{v}(l)\prod_{k=l+1}^{n}\mathbf{A}(k) \right]^{\text{T}} \right\} + \\
\expect\left\{\left[ \displaystyle \sum_{l=0}^{n-1} \mathbf{B}(l)\mathbf{v}(l)\prod_{k=l+1}^{n}\mathbf{A}(k) \right]\left[ \mathbf{B}(n)\mathbf{v}(n) \right]^{\text{T}} \right\} + \\
\expect\left\{\left[ \displaystyle \sum_{l=0}^{n-1} \mathbf{B}(l)\mathbf{v}(l)\prod_{k=l+1}^{n}\mathbf{A}(k) \right]\left[ \displaystyle \sum_{l=0}^{n-1} \mathbf{B}(l)\mathbf{v}(l)\prod_{k=l+1}^{n}\mathbf{A}(k) \right]^{\text{T}} \right\}.
\end{array}   
\label{eq:cov_from_gen_term}
\end{equation}
Owing to assumption A5, it can be shown that the first term in the right side of Eq. \eq{eq:cov_from_gen_term} can be simplified to
\begin{equation}
\mathbf{B}^{2}(n)\mathbf{\Phi}
\label{eq:cov_from_gen_term_1st_term}
\end{equation}
In turn, the second and the third terms in the right side of Eq. \eq{eq:cov_from_gen_term} are zero thanks to assumptions A4 and A5. Next, by expanding the fourth expected value in Eq. \eq{eq:cov_from_gen_term}, it can be noted that all elements containing products of the type $\mathbf{v}(n)\mathbf{v}(m)$ for $n \neq m$ are zero due to assumption A4. By considering only the non-zero elements of the expansion, we get
\begin{equation}
\sum_{l=0}^{n-1}\left[ \mathbf{B}(l) \prod_{k=l+1}^{n} \mathbf{A}(k) \right]^{2}\mathbf{\Phi}.
\label{eq:cov_from_gen_term_4th_term}
\end{equation}
Then, setting the second and third terms in the right side of Eq. \eq{eq:cov_from_gen_term} to zero and using Eqs. \eq{eq:cov_from_gen_term_1st_term} and \eq{eq:cov_from_gen_term_4th_term} in the first and fourth terms, respectively, make Eq. \eq{eq:cov_from_gen_term} equal to \eq{eq:tvar_cov_multivar}.
\subsection*{Determining Eq. \eq{eq:tvar_general_term_form_updated}}

The general-term formula of the TVAR process as function of the defined parameter matrices can be determined from Eq. \eq{eq:general_term_demonstration}. To do so, we use Eqs. \eq{eq:tvar_a_diag_mat} and \eq{eq:tvar_b_diag_mat}, as well as Eqs. \eq{eq:tvar_config1_f0_multivar} to \eq{eq: Gamma_matrix}, to rewrite $\mathbf{A}(n)$ and $\mathbf{B}(n)$ as 
\begin{equation}
\mathbf{A}(n) = \mathbf{P}(n+1)\mathbf{S}(n+1)\mathbf{P}^{-1}(n)\mathbf{S}^{-1}(n)
\label{eq:a_n_expanded}
\end{equation}
\begin{equation}
\begin{array}{l}
\mathbf{B}(n) = \mathbf{\Lambda}\mathbf{P}(n+1)\mathbf{S}(n+1) \times \\
\left[\mathbf{R}(n+1)\mathbf{S}^{-1}(n+1) - \mathbf{R}(n)\mathbf{S}^{-1}(n) \right]^{\frac{1}{2}}.
\end{array}
\label{eq:b_n_expanded}
\end{equation}
Substituting  Eqs. \eq{eq:a_n_expanded} and \eq{eq:b_n_expanded} into Eq. \eq{eq:general_term_demonstration} gives
\begin{equation}
\small
\begin{array}{l}
\mathbf{x}(n+1) = 
\displaystyle\mathbf{x}(0)\prod_{k=0}^{n}\mathbf{P}(k+1)\mathbf{S}(k+1)\mathbf{P}^{-1}(k)\mathbf{S}^{-1}(k) + \\
\displaystyle \sum_{l=0}^{n-1}\mathbf{P}(l+1)\mathbf{S}(l+1) \left[\mathbf{R}(l+1)\mathbf{S}^{-1}(l+1) - \mathbf{R}(l)\mathbf{S}^{-1}(l) \right]^{\frac{1}{2}}\times \\
\mathbf{v}(l)\mathbf{\Lambda}\displaystyle\prod_{k=l+1}^{n}\mathbf{P}(k+1)\mathbf{S}(k+1)\mathbf{P}^{-1}(k)\mathbf{S}^{-1}(k) + \mathbf{v}(n) \times \\
\mathbf{\Lambda}\mathbf{P}(n+1)\mathbf{S}(n+1)\left[\mathbf{R}(n+1)\mathbf{S}^{-1}(n+1) - \mathbf{R}(n)\mathbf{S}^{-1}(n) \right]^{\frac{1}{2}}.
\end{array}   
\label{eq:general_term_demonstration_rearranged}
\end{equation}
In Eq. \eq{eq:general_term_demonstration_rearranged}, the first product can be simplified to
\begin{equation}
\prod_{k=0}^{n}\mathbf{P}(k+1)\mathbf{S}(k+1)\mathbf{P}^{-1}(k)\mathbf{S}^{-1}(k) = \mathbf{P}(n+1)\mathbf{S}(n+1)\mathbf{\Gamma}
\label{eq:prod_simplifiled_a_term}
\end{equation}
due to the fact that $\tilde{\mathbf{x}}(0)=\mathbf{P}^{-1}(0)\mathbf{S}^{-1}(0)\mathbf{x}(0)=\mathbf{\Gamma}\mathbf{x}(0)$ (see Eq. \eq{eq:x_scaled_by_gamma}) and because of cancellations that happen when multiplying consecutive terms. Owing to the latter, the second product in Eq. \eq{eq:general_term_demonstration_rearranged} can also be simplified to
\begin{equation}
\begin{array}{l}
\displaystyle\prod_{k=l+1}^{n}\mathbf{P}(k+1)\mathbf{S}(k+1)\mathbf{P}^{-1}(k)\mathbf{S}^{-1}(k) = \\
\mathbf{P}(n+1)\mathbf{S}(n+1)\mathbf{P}^{-1}(l+1)\mathbf{S}^{-1}(l+1)
\end{array}
\label{eq:prod_simplifiled_b_term}
\end{equation}
By substituting Eqs. \eq{eq:prod_simplifiled_a_term} and \eq{eq:prod_simplifiled_b_term} into Eq. \eq{eq:general_term_demonstration_rearranged} and manipulating and simplifying the resulting expression, it can be shown that Eq. \eq{eq:general_term_demonstration_rearranged} reduce to Eq. \eq{eq:tvar_general_term_form_updated}.

\subsection*{Determinig Eq. \eq{eq:tvar_mean_multivar_updated}}

The expression for the mean vector as function of the TVAR parameter matrices can obtained by simply using Eqs. \eq{eq:a_n_expanded} and \eq{eq:prod_simplifiled_a_term} in Eq. \eq{eq:tvar_mean_multivar}, as well as the scaling defined in Eq. \eq{eq:x_scaled_by_gamma}.

\subsection*{Determing Eq. \eq{eq:tvar_cov_multivar_updated}}

The covariance matrix as function of the TVAR parameter matrices can be computed by substituting Eqs. \eq{eq: P_matrix} to \eq{eq: R1_matrix} into Eqs. \eq{eq:a_n_expanded} and \eq{eq:b_n_expanded}, and the resulting expressions into Eq. \eq{eq:tvar_cov_multivar}. By employing these substitutions and simplifying the obtained expression by using Eq. \eq{eq:prod_simplifiled_a_term}, we get
\begin{equation}
\begin{array}{l}
\mathbf{C}(n) = \mathbf{\Lambda}^2\mathbf{P}^{2}(n+1)\mathbf{S}^{2}(n+1)\times \\
\left\{ \displaystyle\sum_{l=0}^{n} \left[ \mathbf{R}(l+1)\mathbf{S}^{-1}(l+1) \right]^2 - \left[ \mathbf{R}(l)\mathbf{S}^{-1}(l) \right]^2  \right\} \mathbf{\Phi}
\end{array}
\label{eq:cov_tvar_matrix_telescop_1}
\end{equation}
Observe that the right side of Eq. \eq{eq:cov_tvar_matrix_telescop_1} contains a \textit{telescoping sum} with consecutive terms that are cancelled out, yielding
\begin{equation}
\begin{array}{l}
\displaystyle\sum_{l=0}^{n} \left[ \mathbf{R}(l+1)\mathbf{S}^{-1}(l+1) \right]^2 - \left[ \mathbf{R}(l)\mathbf{S}^{-1}(l) \right]^2 = \\
\left[\mathbf{R}(n+1)\mathbf{S}^{-1}(n+1)\right]^2 - \left[\mathbf{R}(0)\mathbf{S}^{-1}(0)\right]^2.
\end{array}
\label{eq:telescoping_sum}
\end{equation}
From Eq. \eq{eq: R2_matrix}, we have $\mathbf{R}(0)=\mathbf{0}$. By using the latter in Eq. \eq{eq:telescoping_sum}, replacing the telescoping sum in Eq. \eq{eq:cov_tvar_matrix_telescop_1}, and simplifying the resulting expression, Eq. \eq{eq:cov_tvar_matrix_telescop_1} becomes Eq. \eq{eq:tvar_cov_multivar_updated}.

\end{document}